\documentclass[conference]{IEEEtran}
\IEEEoverridecommandlockouts
\usepackage[utf8]{inputenc}
\usepackage[T1]{fontenc}
\usepackage{cite}
\usepackage{amsmath,amssymb,amsfonts}
\usepackage{algorithmic}
\usepackage{graphicx}
\usepackage{textcomp}
\usepackage[]{algorithm2e}
\usepackage{xcolor}
\usepackage{url}
\usepackage{breqn}
\usepackage{textcase}
\usepackage[export]{adjustbox}
\usepackage{subcaption}
\usepackage{tabularx}
\def\BibTeX{{\rm B\kern-.05em{\sc i\kern-.025em b}\kern-.08em
    T\kern-.1667em\lower.7ex\hbox{E}\kern-.125emX}}
\begin{document}

\title{Critical concrete scenario generation using scenario-based falsification}
\author{Dhanoop Karunakaran$^{1}$, Julie Stephany Berrio$^{1}$, Stewart Worrall$^{1}$, Eduardo Nebot$^{1}$


\thanks{$^{1}$D.Karunakaran, J. S. Berrip, S. Worrall,  E. Nebot  are with the Australian Centre for Field Robotics (ACFR) at the University of Sydney (NSW, Australia).
       E-mails: {\tt\small \{d.karunakaran, j.berrio, s.worrall,  e.nebot\}@acfr.usyd.edu.au}}
}

\maketitle

\begin{abstract}
Autonomous vehicles have the potential to lower the accident rate when compared to human driving. Moreover, it is the driving force of the automated vehicles' rapid development over the last few years. In the higher Society of Automotive Engineers (SAE) automation level, the vehicle's and passengers' safety responsibility is transferred from the driver to the automated system, so thoroughly validating such a system is essential. Recently, academia and industry have embraced scenario-based evaluation as the complementary approach to road testing, reducing the overall testing effort required. It is essential to determine the system's flaws before deploying it on public roads as there is no safety driver to guarantee the reliability of such a system. This paper proposes a Reinforcement Learning (RL) based scenario-based falsification method to search for a high-risk scenario in a pedestrian crossing traffic situation. We define a scenario as risky when a system under testing (SUT) does not satisfy the requirement. The reward function for our RL approach is based on Intel's Responsibility Sensitive Safety(RSS), Euclidean distance, and distance to a potential collision. Code and videos are available online at \url{https://github.com/dkarunakaran/scenario_based_falsification}.
\end{abstract}
\section{Introduction}

J3016-2018 is an automotive standard from the Society of Automotive Engineers (SAE) that defines the six levels of automation in vehicles\cite{standard2018j3016}. In the higher SAE automation levels, the passengers' and vehicle's safety accountability transitions from the human driver to the autonomous system. So, evaluating the SUT thoroughly before deploying it on public roads is essential. Traditional testing methods are not suitable for evaluating the Highly Automated Vehicle (HAV) as many challenges such as non-determinism in testing arise due to the ML and probabilistic algorithms\cite{koopman2016challenges}.

\begin{figure}[t]
\includegraphics[width=0.98\columnwidth]{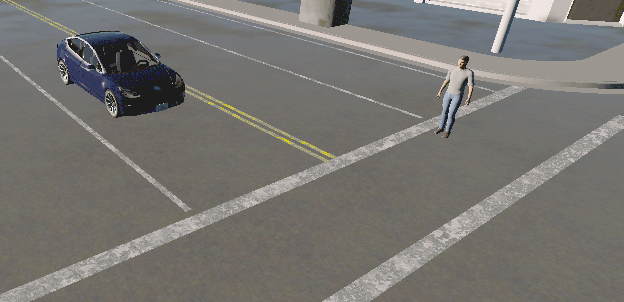}
\caption{\small Experiment in a CARLA simulation: a pedestrian crossing the street while the vehicle is driving. }
\label{fig:ped_cross_sample}
\end{figure}

Road testing is a common approach to evaluating HAV. However, the majority of the driving task is low risk and of little value for testing\cite{xinxin2020csg}. Scenario-based testing is a promising approach as it focuses only on meaningful scenarios so that it can reduce the test effort required in safety evaluation \cite{putz2017system, elrofai2016scenario, amersbach2019functional, ponn2020identification}.
There are three abstraction levels for a scenario representation:  functional, logical, and concrete scenario \cite{menzel2018scenarios}. A functional scenario is at the top of the abstraction level and is often represented by a linguistic description of a scenario. The logical scenario defines different parameters along with their range/distribution. Finally, the concrete scenario specifies a fixed value from the range/distribution for each of the parameters. This representation is used to describe test cases to evaluate the SUT. In this paper, a risky or challenging scenario is referred to as a dangerous concrete scenario.

There are two types of techniques in scenario-based safety evaluation: testing and falsification \cite{riedmaier2020survey}. Testing determines whether the SUT satisfies the safety requirement by assessing the system over a finite set of concrete scenarios. The falsification approach finds the system's weakness by searching for a critical scenario in which SUT violates the safety requirement. Even though falsification does not create statistical safety arguments, it is useful for identifying the systems' weaknesses. This approach can complement the testing techniques for a more reliable safety assessment of Highly Automated Vehicles.   

This paper proposes a Reinforcement Learning (RL) based scenario-based falsification method to search for critical scenarios in a pedestrian crossing. This technique aims to search for a critical concrete scenario that have a maximum reward in the explored space. In other words, the method searches for the pedestrian trajectory that result in high-risk scenario or even collision with a Vehicle Under Test (VUT). The method will encounter many dangerous scenarios during the overall process, but it learns towards a problematic scenario that can gain maximum reward in the explored space through the reward function. Our method focuses on identifying the parameter settings of a concrete scenario at the beginning of each episode. In contrast, other RL-based approaches\cite{koren2018adaptive,liu2021reinforcement} in the literature concentrate on identifying the parameter settings at each time step to create a detailed trajectory.

In this experiment, we create the reward function for RL using Intel's Responsibility-Sensitive Safety (RSS) metric \cite{mobileye_rss}, the Euclidean distance of pedestrian with respect to the ego-vehicle, and a cost that detects a collision. The pedestrian crossing illustrated in Figure \ref{fig:ped_cross_sample} is the selected traffic situation to test our method. We have used five parameters, each with an average number of twelve values. Even though we have chosen only five parameters to represent the traffic situation's search space/parameter space, it can create approximately $12^5 = 248832$ possible concrete scenarios. Adding more parameters increases the search space exponentially. The brute force is an exhaustive search that usually takes many iterations to find the critical scenario in an ample search space. This leads to spending more time for testing a production vehicle. In \cite{zhao2016accelerated}, authors argued that Random search methods such as Crude Monte Carlo (CMC) take longer to find rare scenarios or edge cases. Thus, optimisation algorithms such as Reinforcement Learning (RL) are suitable to achieve the desired outcome.  

In a pedestrian crossing, a critical scenario is a collision between a vehicle and a pedestrian. As shown in Equation~\ref{eqn:stl_safety_requirement}, we have defined the safety requirement in Signal temporal logic (STL). 
STL is mainly used for cyber-physical systems (CPS) such as AV, and we use the STL requirement from the work of \cite{tuncali2019requirements}.

\begin{equation}
\label{eqn:stl_safety_requirement}
req = \square(\lnot \pi_{i,clash})
\end{equation}
where $\pi_{i,clash} = dist(i, ego) < \epsilon_{dist}$.

The requirement states that the ego vehicle should not collide with any traffic participants. $i$ indicates a traffic participant such as a vehicle and pedestrian that share the environment with the ego vehicle. The $dist(i, ego)$ gives the Euclidean distance between traffic participant, $i$ and ego-vehicle. $\epsilon_{dist}$ is a minimum safe distance between two parties required to avoid accidents. 

The proposed method aims to search and find a challenging concrete scenario in which SUT violates the safety requirement defined in Equation~\ref{eqn:stl_safety_requirement}. The result shows that our RL-based approach converges to a critical scenario with maximum reward in the explored parameter space. In this paper,  we are interested in evaluating the SUT by finding a critical concrete scenario specific to the system. It may or may not be critical to other systems.


\section{Related work}

In this section, we present the existing literature related to scenario-based falsification for assessing Automated Vehicles.

The authors in \cite{de2017assessment} parameterise the real-world scenarios and then extract those scenario parameters to store them in a database. In the next step, they fit a distribution to parameters for creating scenarios by sampling from the fitted distribution. The generation of scenarios utilises Monte-Carlo simulation and Importance Sampling. However, Monte-Carlo is a random search and cannot guarantee finding edge cases even when employing Importance Sampling to create more critical scenarios.
In \cite{zhao2017accelerated}, Zhao Et.al. build a statistical model of the SUT's behaviour by processing the collected real-world data. This work is focused on creating the model in a  lane-following traffic situation. As most real-world data is not challenging, they skewed the model to create problematic scenarios. Nevertheless, one drawback of the model-based approach is that the result is inaccurate if the model approximation to real-world behaviour is incorrect. 

Bayesian optimisation (BO) can be employed to optimise expensive-to-evaluate functions or systems. The core idea in BO is the use of Bayes rule to train the model to generate challenging scenarios. The scenario is created with parameters '$X$' and '$Y$' representing input parameters and output, respectively. Before the optimisation starts, a few data points are created by randomly selecting input parameters $X$, from the search space and executing them in a simulation. The model is initially fitted with random data points. In the iteration stage, the acquisition function samples the search space and selects the best $X$ based on the fitted model's prediction. In the next step, the simulation runs the generated scenario using input parameters provided by the acquisition function and produces the actual output $Y$. These results are added to the data points used to fit the model. This process repeats until the end criteria is reached. In \cite{gangopadhyay2019identification}, the authors introduced the BO to find challenging scenarios for evaluating HAVs. They use the learned model to find the challenging scenarios once the optimisation process has found an approximation of the model. However, BO has a theoretical limit on the number of parameters it could be used. 

In \cite{althoff2018automatic} Althoff et al. proposed an evolutionary algorithm (EA) to search for critical concrete scenarios. The method enables searching for the optimal global solution (critical concrete scenario) with respect to a fitness function. The algorithm follows the process of natural selection, where the fittest individuals are selected to generate a new generation of individuals. The fitness function checks how fit the chosen individuals are. Similarly, the function identifies the parent scenarios to create a new scenario. This process repeats until a termination criteria is reached. Even though it can generate diverse scenarios, sampling the edge cases from the learned generative models remains unsolved since the edge cases have a low probability of occurrence \cite{ding2020learning}.

The underlying principle of combinatorial testing is that most failures are the result of the interaction between a small number of parameters\cite{kuhn2013introduction}. The study from the \cite{kuhn2004software} argued that there are six or fewer parameters that caused the majority of the known failures. The most common type of combinatorial approach is pairwise testing, which relies on the argument that the interaction of two parameters causes many failures. \cite{tuncali2019search} uses combinatorial testing to evaluate the Deep Learning (DL) based object detection algorithm. The author argued that DL-based systems are sensitive to variations of a few parameters, and combinatorial testing is best suited for evaluating such a system.
In \cite{felbinger2019comparing}, the author introduced combinatorial testing to find the critical scenario for assessing the Autonomous Emergency Braking (AEB) System. The challenge with this approach is selecting the salient parameters, which in most cases are hard to identify.

A model-free approach to creating challenging scenarios in the cyber-physical system using simulated annealing is proposed in \cite{aerts2018temporal}. The simulated annealing process comes from metallurgy, where heating and controlled cooling are used to reduce defects. Similarly, the temperature parameter controls the exploration and exploitation of the optimisation. During the initial stage, the temperature value will be high, enabling the exploration of various scenarios. Towards the end, the temperature value decreases, reducing the exploration and moving towards the exploitation stage. It means that the method started to choose the scenarios as possible critical scenarios. The primary concerns with this approach are that the algorithm may take a long time to complete.  

Some work in the literature uses Reinforcement Learning (RL) to generate challenging scenarios for the assessment of Autonomous Vehicles. In \cite{koren2018adaptive}, Koren presents an RL-based approach to find the most likely failure scenarios. The method enables the learning of pedestrian trajectories that end up in a collision with the SUT. Mainly, the experiment has three parameters: acceleration of the pedestrian and two noise parameters representing the uncertainty in the SUT's estimated position and velocity. The result shows that their method learns the parameter values required at each timestep to create a collision trajectory. It indicates that the SUT fails to avoid the collision with a pedestrian and requires further design changes before releasing it on public roads. \cite{liu2021reinforcement} uses Reinforcement Learning to search for collision scenarios in an intersection. They use parameters for the ego-vehicle (speed and position) and other vehicles (speed, position, and heading) to create dangerous trajectories. Similar to \cite{koren2018adaptive}, their method learns what parameter values are required at each timestep to create challenging scenarios. The previously mentioned RL approaches do not include the parameters such as weather with the different presets. The main reason is that these parameters do not vary in each timestep.

Neural architecture search (NAS) is a hyperparameter optimisation approach to automate the design of a neural network for a specific task \cite{nas}. The authors applied the policy gradient RL method called REINFORCE to teach the method to predict the hyperparameters of the Neural Network for a defined task. We apply this method to the falsification problem. In our work, the parameters are used to generate a specific scenario. A RL algorithm learns the combination of parameters that correspond to a critical concrete scenario. In \cite{xu2020worst},  Xu et al. uses NAS to search for the worst perception scenarios using a vehicle detection algorithm within three publicly available perception datasets. The main focus of their approach is to find critical perception scenarios. Our work also applies NAS to search for critical scenarios but concentrates on events in which the behaviour of the SUT does not meet the defined safety requirement. We use an autonomous emergency braking (AEB) driving function as SUT, which combines perception and decision-making components. Then focuses more on the whole SUT's behavioural aspect instead of one part of the entire driving function.

\section{Background}

This section introduces the concepts of reinforcement learning (reinforce), CARLA simulator and responsibility sensitive safety metric used in proposed method presented in this paper.

\subsection{REINFORCE}
A reinforcement learning (RL) algorithm finds an optimal policy that maximizes a reward by interacting with the environment that is modelled as a Markov Decision Process (MDP) \cite{sutton2018reinforcement}. In MDP, all states satisfy the Markov property, referring to the fact that the future depends on the current state only \cite{sutton2018reinforcement}. We can implement Reinforcement Learning(RL) in three ways: dynamic programming, Monte-Carlo Methods, and Temporal-Difference Learning. REINFORCE is a Monte-Carlo method \cite{levine2017cs}. An agent applies an action to the current state and will then navigate to the goal state to compute the next reward. In contrast, Temporal Difference Learning and Dynamic programming use bootstrapping techniques to speed up the learning. REINFORCE is a policy gradient algorithm that directly manipulates the policy to reach the optimal policy that maximises the expected return. If we are using Neural Network (NN) as the agent or controller, then the weights of the NN are the policy. The learning process enables the weights' update to find the optimal policy that predicts the desired action given a state. The pseudo-code for the REINFORCE \cite{levine2017cs} is shown is shown in Algorithm~\ref{algo_reinforce}. 


\begin{algorithm}[h]
 \KwResult{Optimal policy $\pi_{\theta}$}
 \While{optimal policy $\pi_{\theta}$ is not found}{

1. Sample N trajectories using policy $\pi_{\theta}$\;
2. Evaluate the gradient of the objective function J using the below expression:
\begin{equation} 
\label{eqn:reinforce1}
\nabla J(\theta) \approx \frac{1}{N}\sum_{\tau\in N}\sum_{t=0}^{T-1}\nabla_\theta log \pi_{\theta}(a_t, s_t) R(\tau)
\end{equation}
 Where N is the number of trajectories and R is the total return of a trajectory\;
3. Update the policy parameters
\begin{equation} 
\label{eqn:reinforce2}
\theta = \theta+\alpha \nabla J(\theta)
\end{equation}
 }
 \caption{Pseudocode for the REINFORCE}
 \label{algo_reinforce}
\end{algorithm}

We use the cross-entropy loss representing the grad-log policy in the policy gradient expression. It opens up the use of NN as the functional approximator to describe the stochastic policy $\pi_{\theta}$.

\subsection{Responsibility Sensitive Safety (RSS)}

Intel's Mobileye proposed a metric that computes the minimum safe distance required to keep the vehicle safe \cite{mobileye_rss}. The ultimate goal of RSS as a formal method is to guarantee that an agent will not cause an accident rather than to ensure that an agent will not be involved in an accident. We can compute RSS longitudinally and laterally based on the proposed formulas. In this paper, we focus on the safe longitudinal distance. By definition, the safe longitudinal distance $d_{min}$ is the minimum distance required for the ego vehicle to stop in time if a vehicle in front brakes abruptly.

\begin{equation} \label{eq1}
d_{min}=\left[v_r\rho+\frac{1}{2}a_{max,a}\rho^2+\frac{(v_r+\rho a_{max,a})^2}{2a_{min,b}}-\frac{v_f^2}{2a_{max,b}}\right]
\end{equation}

where $v_r$  and $v_f $ are the velocity of the agent vehicle and front vehicle, respectively, $a_{min,b}$ is the minimum reasonable braking force of the agent vehicle, $a_{max ,b}$ is the maximum braking force of the front vehicle, $a_{max,a} $ is the maximum acceleration of the front vehicle, and $\rho$ is the agent vehicle response time.

In this work, we use RSS as one of the metrics to compute the reward function that enables the RL agent's learning.

\subsection{CARLA}


CARLA is an open-source simulator for developing, training, and validating autonomous urban driving systems\cite{carla_simulator}. It is a commonly used simulation tool for academic research based on Unreal Engine to run the simulation. The simulator provides Python and C++ APIs to control it. This tool allows the complete control of digital assets such as static and dynamic actors to create scenarios. CARLA supports many sensors available in the Highly Automated Vehicle, making it suitable for developing autonomous systems. Furthermore, the ROS bridge provided by CARLA enables the integration of the simulator with any ROS-based systems.

\section{Scenario-based falsification method}

\begin{figure}[h]
\centering
\includegraphics[width=0.95\columnwidth]{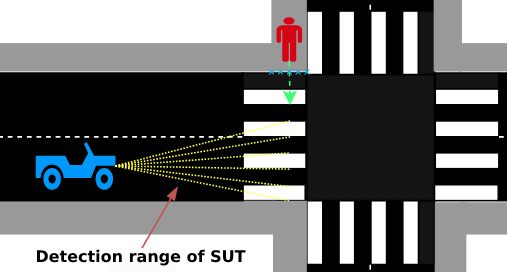}
\caption{\small Pedestrian crossing intersection}
\label{fig:pedestrian_crossing}
\end{figure}

Our proposed method searches for a problematic concrete scenario in a pedestrian crossing. Figure~\ref{fig:pedestrian_crossing} illustrates the pedestrian crossing example. The method enables the learning towards the critical scenario where the SUT fails to meet the safety requirement. The following two subsections explain various components of the falsification method and implementation details.

\subsection{Components}

\begin{figure}[h]
\vspace{5mm}
\includegraphics[width=0.98\columnwidth]{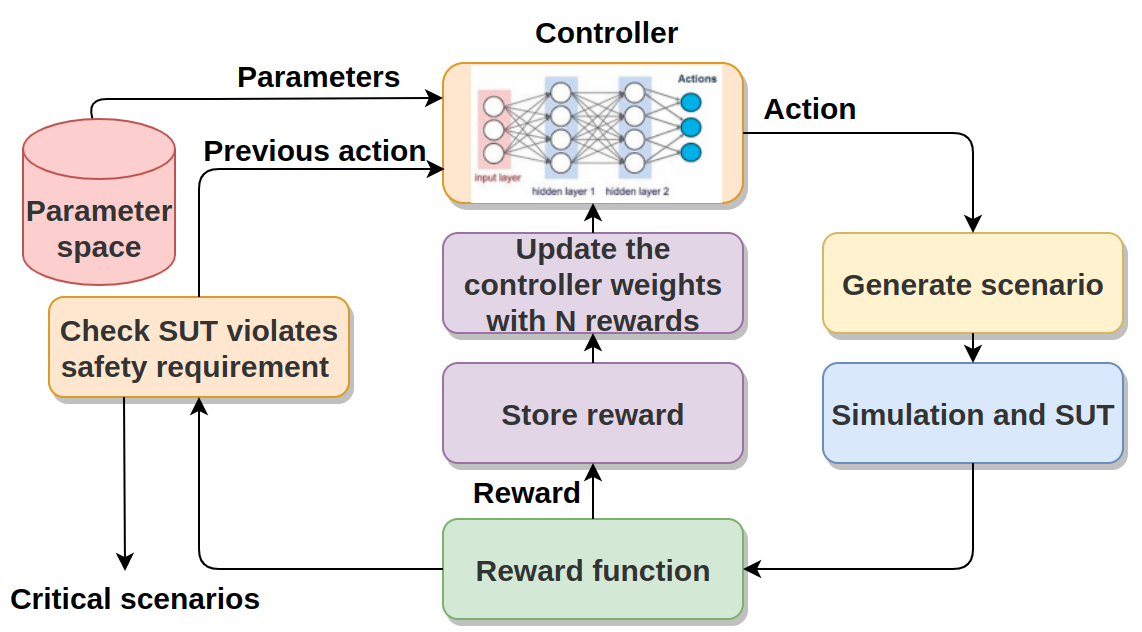}
\caption{\small This figure illustrates in detail the architecture of the proposed method. The aim is to find challenging scenarios with the maximum return by optimising the controller. Each action produced by the controller is a combination of parameter values taken from the search space, finally the weights of the controller are updated by the reward.}
\label{fig:architecture}
\end{figure}

Figure \ref{fig:architecture} depicts the architecture of the proposed method. The detailed description of each component is explained below.

\subsubsection{System Under Test (SUT)}
In this paper we use CARLA's emergency braking system as SUT. The system controls the vehicle's speed to avoid a collision when it detects an object within its field of view. Safety evaluation against the requirement is done at the end of the development lifecycle in the traditional engineering process. At this stage of the process, a test engineer may not know the internal specifications of the system. Similarly, the actual system specification is unknown if a third party performs the evaluation. So, it is essential to develop safety assessment methodologies that treat the SUT as a black-box system.
\begin{figure}[h]
\includegraphics[width=0.95\columnwidth]{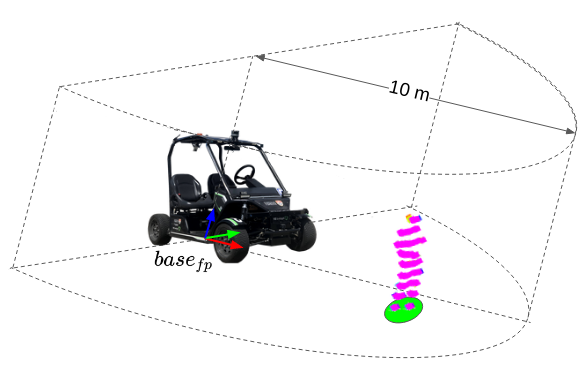}
\caption{\small Detection range of CARLA's emergency braking system}
\label{fig:sut}
\end{figure}

\subsubsection{Parameter space}
This work uses parameter space as an alternative term for the logical scenario. It contains the frequency distribution of each parameter. We have considered five parameters in this experiment: offset from ego-vehicle's original position  (\textit{ego-offset-pos}), pedestrian acceleration (\textit{ped-accel}), pedestrian velocity (\textit{ped-vel}), offset from the pedestrian orginal position (\textit{ped-offset-pos}), and weather. The list of parameters and their distribution is shown in Table~\ref{table:search_space}.
The CARLA simulator places the ego-vehicle and pedestrian in a user defined position. In order to change the longitudinal position of the ego-vehicle and pedestrian, we use the offset parameter. Carla has 15 weather settings, and we use 10 out of these values in the weather parameter. Each value represents the weather preset. For instance, zero points to the 'Clear Noon', whereas fourteen indicates 'Wet Sunset'.

The number of samples is an indicator of how many values are sampled from the distribution of each parameter.

\begin{table}[h!]
\centering
 \caption{List of simulation parameters}
 \label{table:search_space}
 \begin{tabular}{|c c c|} 
 \hline
 Parameter & Distribution & No. of samples \\ [0.5ex] 
 \hline\hline
 \textit{ego-offset-pos} & $\mathcal{U}(1,10) m$ & 10\\ 
 \textit{ped-accel} & $\mathcal{U}(0,0.1) m/s^2$ & 10\\
 \textit{ped-vel}  & $\mathcal{N}(1.46,\,0.24) m/s$ & 25\\
 \textit{ped-offset-pos} & $\mathcal{U}(3,4.5) m$ & 4\\
 weather & $\mathcal{U}(0,14)$ & 10\\ [1ex] 
 \hline
 \end{tabular}
\end{table}







\subsubsection{State space and action space}
The proposed method aims to generate the best parameter combination that provides the highest return. The \textit{action} contains the selected parameter combination to create a scenario in simulation. The method learns how to optimise the \textit{action} so it contains the best parameter combination. In RL, the agent provides the \textit{action} given a state, but this experiment has no separate states; instead, the previous action is given as the current state. The reason for it is we only need to learn the right parameter combination, which does not depend on any state.

There are five parameters in the search space, so action contains the value taken from each parameter.  So, action is a list of 5 parameter values.

\subsubsection{Reward function}
A reward function gives the score based on the state of the environment once the agent applies the action. We compute three types of rewards in this experiment. Firstly, we measure how many timesteps the Euclidean distance between the vehicle and the pedestrian is less than the RSS distance. Such timesteps are classified as high-risk, and we use notation, $highrisk_{ts}^{RSS}$, to represent the total risky timesteps. We noted in the experiment that the higher the number of risky timesteps, the more likely it is to have a collision. So, this type of reward encourages the RL to search for scenarios that produce more high-risk timesteps. We normalised the reward to a small value range $new_x$ as we noted that a large value could affect the controller's ability to learn.

\begin{equation}
\label{eqn:normalize2}
new_x = (b-a)*\frac{x-min}{max-min}+a
\end{equation}


Where $x$ is the total number of high-risk timesteps, the minimum (min) value is zero in this case, and the maximum(max) value is the total no. of timesteps in an episode. The minimum value $a$ is -0.01 and the maximum value $b$ is 0.01. These values were found through experimentation, where a lower reward produced an extensive exploration without converging quickly into the local maxima/minima.

The second type of reward is based on the Euclidean distance between pedestrian and ego-vehicle. At the end of every episode, we calculate the final distance. When the distance gets smaller, it is more likely to have a potential collision. So we compute the higher reward for smaller distances and the lesser for longer distances. The value of the reward ranges between -0.01 and 0.01. 

The collision is the ultimate scenario we need to find, and rewarding the higher value for such scenarios can teach the controller to bias toward the problematic scenarios. We wanted a higher value than 0.01, which is the highest reward for other metrics. Also, a reasonable lesser reward such as a way it can promote learning and avoid the quickly stuck into the local minima. We choose 0.25 as the reward for collision through the trial and error method.

Equation~\ref{eqn:reward} shows the reward function with all three types of rewards, and the output of the reward function will be the sum of all three types.  

\begin{equation}
\label{eqn:reward}
R = \Bigg\{\begin{matrix}-0.01\ to\ 0.01&\ highrisk_{ts}^{RSS}\\
-0.01\ to\ 0.01&\ Euclidean\ distance\\
0.25&\ collision\\
\end{matrix}
\end{equation}

\subsubsection{Controller}
The main task of the controller is to optimise the action so that it generates a critical scenario with a maximum reward. The overall architecture enables the learning using a policy gradient RL algorithm called REINFORCE. The controller's goal is to learn the policy that maximises the return in RL terms. The controller in this paper is a Recurrent Neural Network (RNN), and learning is enabled by updating its weights. We use the rewards to update the weights of the RNN, in this process the RNN learns the optimal policy capable to generate the critical scenario with a high reward.

\begin{figure}[h]
\includegraphics[width=0.95\columnwidth]{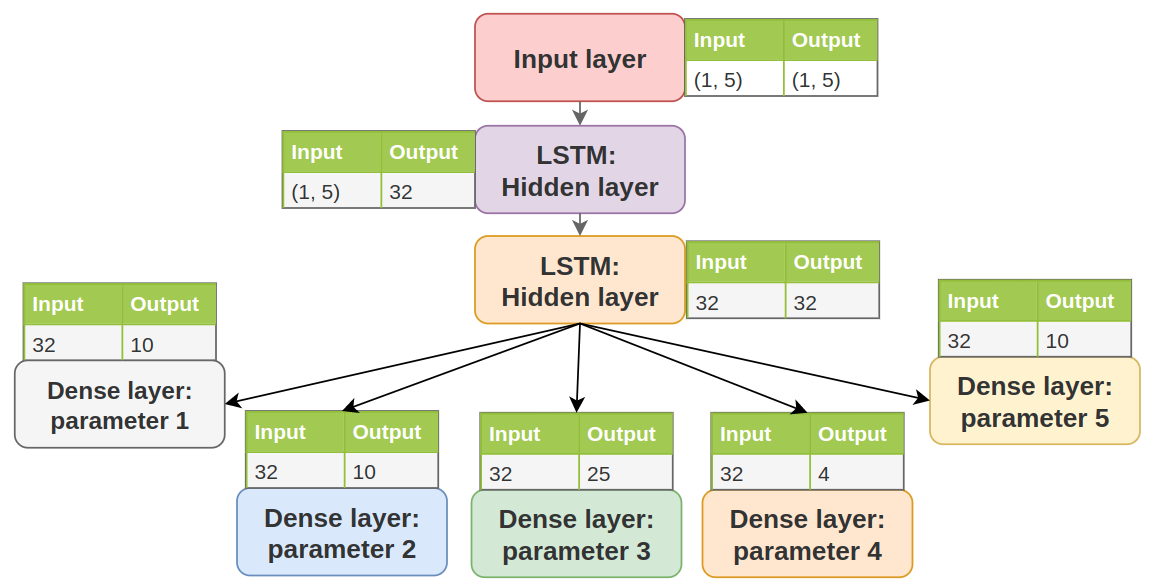}
\caption{\small The plot depicts the architecture of the controller. It has a input layer, two hidden LSTM layers, and a final layer with 5 output. Each output layer represents a different parameter.}
\label{fig:controller_architecture}
\end{figure} 

As shown in Figure \ref{fig:controller_architecture}, the controller has an input layer, followed by two hidden layers, and finally, the output layer contains five units. The whole architecture is LSTM based, so the input layer expects sequential data. 
We sequentially provide previous action data; an example is shown in Equation~\ref{eqn:action_sample}.The output layer has an output unit for each parameter. We would need to increase the number of output units according to the parameter count. 

\begin{equation}
\label{eqn:action_sample}
action, a = [3\ 0.051\ 1.178\ 3.5\ 8]
\end{equation}

The Neural Network (NN) training process in REINFORCE is slightly different from standard NN applied in other tasks. In standard NN, one compute the gradient of the loss function at every epoch and backpropagate the gradient to update the network weights. In this case, we need the ground truth in a supervised learning setting to compute the proper gradient using the loss function. In contrast, model-free RL is a trial and error method and does not have any ground truth to compute the gradient. The reward works as an indicator of how good the predicted output is. The REINFORCE is a model-free RL algorithm where the reward is multiplied by the gradient \textit{log} of NN's output as shown in Equation~\ref{eqn:reinforce1}. Based on Equation~\ref{eqn:reinforce2}, we update the weights of the NN, $\theta$, with the gradient expression computed in Equation~\ref{eqn:reinforce1}. In this way, we can teach the controller to learn toward the maximum reward.

We choose this value to balance exploration versus exploitation. In each update, our method learns how to generate critical scenarios. If we have a value lower than 25, the method learns fast, reducing exploration. In contrast, higher values increase the exploration but take more time to converge.

\subsection{Implementation}
Initially, the parameter space is created by sampling from the distribution shown in Table~\ref{table:search_space}. The generated action contains five parameter values. The method uses these values to construct a scenario in the CARLA simulator. In this work, we focus on developing a critical case (collision with a pedestrian) in a pedestrian crossing.

During the initial stage, the controller produces actions randomly. A later stages, the controller learns the optimal policy to create an action that leads to a challenging/risky scenario for the SUT. The final reward is computed at the end of each episode based on the reward function. A high reward applies to the action that leads to a failure (i.e. collision), and a low reward denotes less risky scenarios.

We have set the controller to operate for 4000 episodes in this work. This number depends on the amount of parameters and possible values they can take.

\section{Results}

Our method searches for a problematic scenario by biasing the learning toward such a scenario in which the emergency braking system violates the safety requirement. We use 'do not collide' as the safety requirement since collision is a dangerous scenario. We set up the reward function so that common scenario will have low rewards compared to the high rewards for risky scenarios. As discussed in the controller and REINFORCE section, the highest reward can influence the controller to learn how to generate potential collision scenarios.

\begin{figure}[h]
\vspace{5mm}
\centering
\includegraphics[width=0.99\columnwidth]{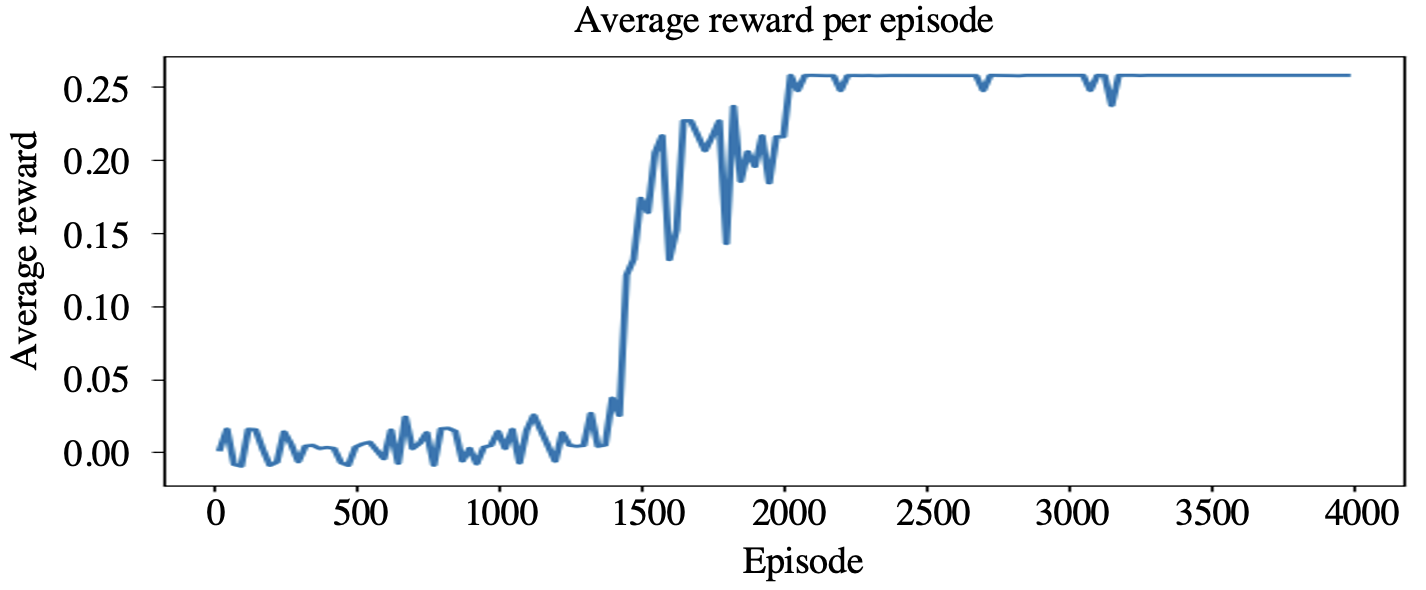}
\caption{\small Average reward per episode, showing the overall learning of the controller. After the 1500$^{th}$ episode, the approach learned the action to obtain a maximum return in explored space. It indicates that the controller has started to predict an action that maximises the return in explored space around the 1500$^{th}$ episode. The reward is almost constant from the 2000$^{th}$ episode onwards. At this stage, generated actions are similar.}
\label{fig:reward_per_episode}
\end{figure}

\begin{figure*}[t]
\vspace{5mm}
    \centering
    \begin{subfigure}[b]{0.98\columnwidth}
         \centering
         \includegraphics[width=0.99\columnwidth]{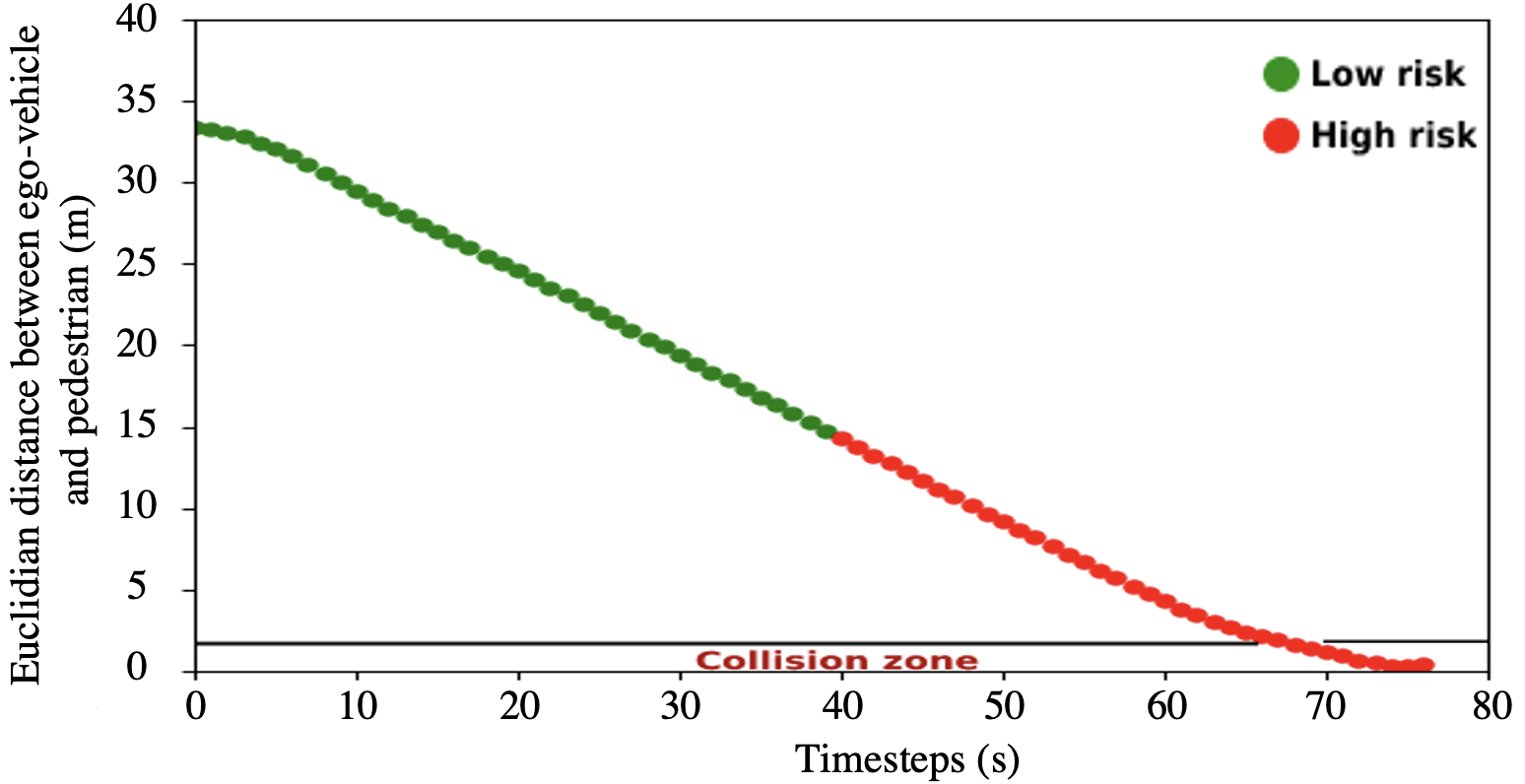}
         \caption{Challenging scenario}
         \label{fig:challenging_scenario}
    \end{subfigure}
    \hspace{3mm}
    \begin{subfigure}[b]{0.98\columnwidth}
         \centering
         \includegraphics[width=0.99\columnwidth]{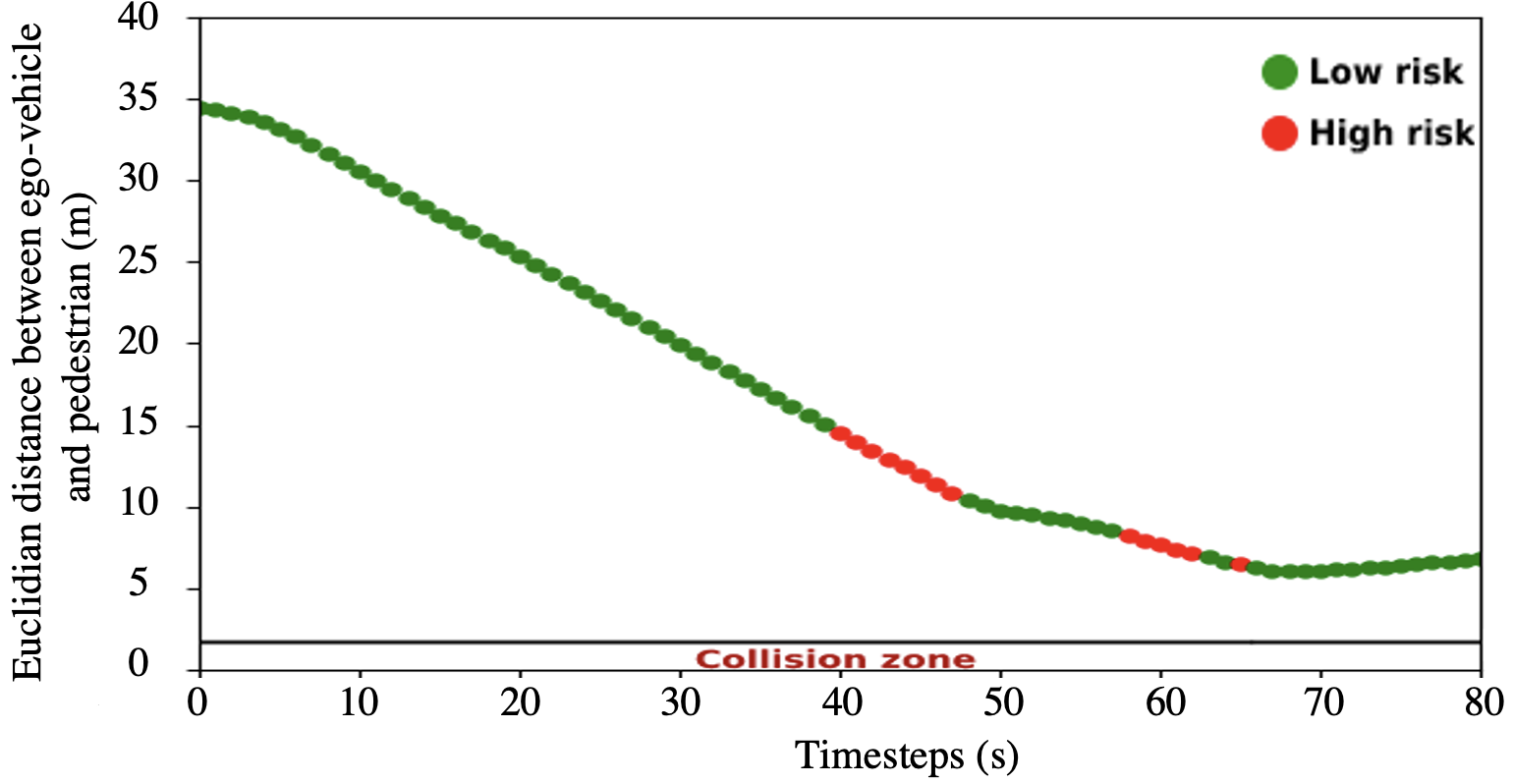}
         \caption{Non-challenging scenario}
         \label{fig:success_scenario} 
    \end{subfigure}
    \caption{\small The plot shows the Euclidean distance between pedestrian and ego-vehicle over timesteps for challenging and non-challenging scenarios. Euclidean distance closer to zero suggests that traffic participants end up in a collision. We mark all the areas under less than one meter Euclidean distance as collision zone. Figure \ref{fig:challenging_scenario} shows the challenging scenario learned by the method that has the highest reward. In this scenario, the final Euclidean distance finishes in a collision zone. It also depicts a more significant number of high-risk timesteps towards the end, indicating many high-risk timesteps can lead to collision scenarios. Figure \ref{fig:success_scenario} explains the typical non-challenging scenario in the experiment. The Euclidean distance does not touch the collision zone and implies that the vehicle stopped before hitting the pedestrian. The figure also shows the safe scenario has a fewer high-risk timestep.}
\label{fig:scenarios_two}
\end{figure*}
As shown in Figure \ref{fig:reward_per_episode}, the proposed method begins to converge after the 1500$^{th}$ episode. The average reward remains almost constant from the 2000$^{th}$ episode onwards. The controller learned to predict the correct action, creating challenging cases with maximised reward. As shown in the figure, ultimately, it converges into a critical scenario with maximum reward in explored space. 



\begin{table}[h!]
\centering
 \caption{parameter values for a challenging and non-challenging scenario}
 \label{table:learned_parameter}
 \begin{tabular}{|c c c|} 
 \hline
 Parameter & Challenging & Non-challenging \\ [0.5ex] 
 \hline\hline
 \textit{ego-offset-pos} & 9.0 $m$ & 7.0 $m$\\ 
 \textit{ped-accel} & 0.007 $m/s^2$ & 0.076 $m/s^2$\\
 \textit{ped-vel}  & 1.237 $m/s$ & 1.808 $m/s$\\
 \textit{ped-offset-pos} & 3.5 $m$ & 3.5 $m$\\
 weather & 8.0 & 8.0\\ [1ex] 
 \hline
 \end{tabular}
\end{table}

The challenging scenario learned by the proposed method is depicted in Figure \ref{fig:challenging_scenario}, and corresponding learned parameter values are shown in Table \ref{table:learned_parameter}. 
The diagram illustrates the Euclidean distance between the ego-vehicle and pedestrian. In the figure, we mark the area under one meter distance as a collision zone. If the Euclidean distance is closer to zero, it implies that the ego-vehicle ends up in a collision with the pedestrian. Figure~\ref{fig:challenging_scenario} is a challenging scenario as the Euclidean distance finishes at the collision zone. We also show high-risk and low-risk timesteps in red and green, respectively. This work defines a timestep as high-risk when the Euclidean distance is lower than the minimum safe distance computed by RSS. Other timesteps are called low-risk timesteps. The plot shows that the last part of the scenario has numerous high-risk timesteps, indicating the vehicle did not reduce the speed required to avoid the collision while the pedestrian was crossing. The reason for the crash is that vehicle wasn't aware of the pedestrian crossing throughout the scenario due to the short field of view of the perception system used by the emergency braking system. This example shows that method finds the scenario in which the emergency braking system violates the safety requirement as defined in Equation ~\ref{eqn:stl_safety_requirement}.

Figure \ref{fig:success_scenario} and  parameter values in the non-challenging section of Table \ref{table:learned_parameter} illustrates a non-challenging scenario. 
The Euclidean distance finishes six-meter ahead of the collision zone, stating the vehicle sees the pedestrian at the crossing and stops before hitting. All the generated scenarios have similar initial ego-vehicle speeds, and the speed variation depends on whether the SUT detects the pedestrian at the later stages. However, these variations are not huge, so it is expected to have a few high-risk timesteps in the non-challenging scenarios compared to the risky scenarios. In this case, a pedestrian walked with higher acceleration than the pedestrian in the challenging scenario. As a result, the speed of the pedestrian is higher and reaches the centre of the road long before the vehicle arrives at the crossing area. It made the pedestrian in the field of view of the vehicle, and the emergency braking system activated to stop the vehicle before hitting the pedestrian. As we can see in Table~\ref{table:learned_parameter}, the velocity of the pedestrian in the non-challenging scenario is higher than the problematic scenario that puts the pedestrian in the field of view of the vehicle.

\begin{figure}[h]
\centering
      \includegraphics[width=0.95\columnwidth]{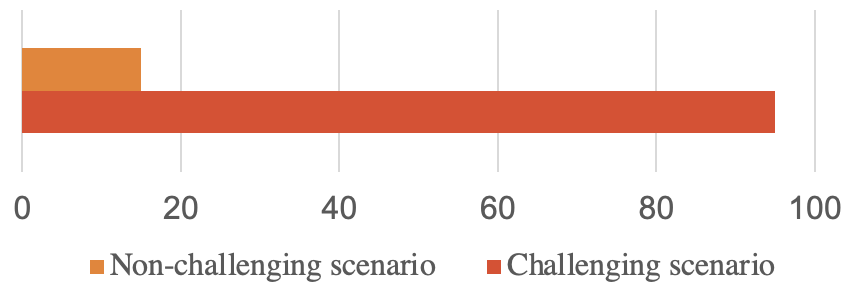} 
      \caption{\small The plot illustrates the percentage of high-risk timesteps for  challenging and non-challenging scenario as shown in Figure \ref{fig:scenarios_two}. It is based on the percentage of time where the system displays high-risk behaviour. The plot shows that the problematic scenario has more high-risk timesteps than the non-challenging scenario. We assume that risky time steps in the scenario's initial period may not be as impactful as the last part. Based on that assumption, we have calculated the high-risk percentage from the second half.}
      \label{fig:bar_chart}
\end{figure} 

We assume that high risk timesteps at the initial stage of the scenario may not have as much impact as in the later stage. Because of this, we calculate the high-risk percentage from the second half of the scenario. The bar chart in Figure \ref{fig:bar_chart} shows that over $95\%$ of the time, SUT exhibits high-risk behaviour in a challenging scenario(figure \ref{fig:challenging_scenario}), whereas it is less than $15\%$ in the non-challenging scenario. As shown in the collision scenario example, more disagreement with RSS and distance between both participants can lead to collision.

\begin{figure}[h]
\centering
      \includegraphics[width=0.93\columnwidth]{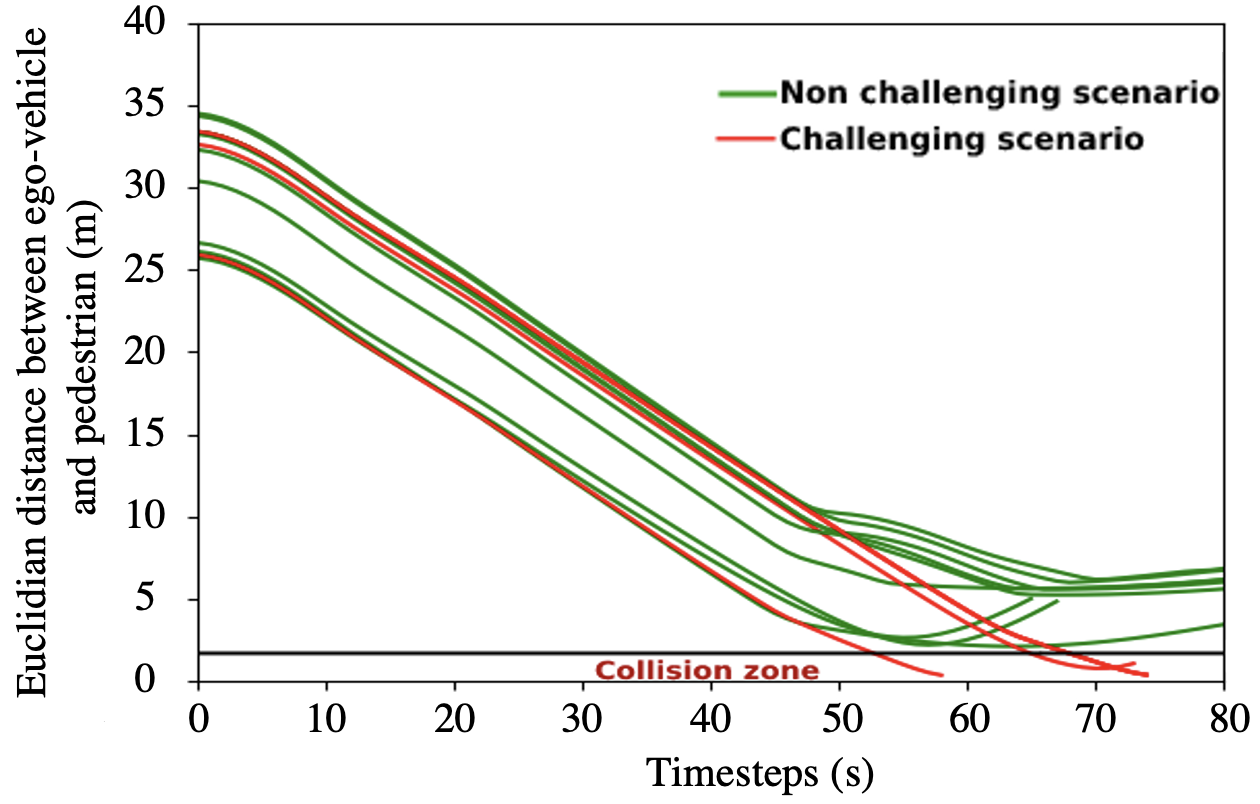} 
      \caption{\small The plot shows the Euclidean distance between vehicle and pedestrian for 14 scenarios. The red coloured lines indicate collision scenarios, and the green colour indicates the non-challenging scenarios.}
      \label{fig:all_scenarios}
\end{figure} 

In Figure \ref{fig:all_scenarios}, we have selected 14 scenarios to show their Euclidean distance over time and which three of those scenarios end up in a collision zone. During the learning process, the method encounters many dangerous scenarios, and in the end, it optimises toward a risky scenario. We have shown three critical scenarios in this plot, represented in red and 11 non-challenging scenarios in green colour.

\section{Discussion and Conclusion}


This paper presents a scenario-based falsification method to generate a critical concrete scenario to evaluate a system under test. We have defined the safety requirement to find a concrete scenario in which SUT violates them. The result shows that the method can search and detect such critical scenarios. The method is not deterministic because the values of the parameters are explored randomly in the early stage. Due to the initial random search, each experiment has a different explored space. So, the values are not repeatable in the subsequent experiment. However, it allows us to find diverse parameter combinations for collision scenarios within a parameter space. 

The parameter space is built in this work based on expert knowledge. But, it is essential to represent the parameter space with real-world traffic participant behaviour. In the extended work, we are building lane-change scenario parameter space from real-world data and use this method to find critical concrete scenarios from it

 \section*{Acknowledgment}
 This  work  has  been  funded  by  the  Australian  Centre  for Field Robotics (ACFR), University of Sydney and Insurance Australia Group (IAG) and iMOVE CRC and supported by the  Cooperative  Research  Centres  program,  an  Australian Government initiative.

\bibliographystyle{IEEEtran}
\bibliography{main.bib}

\begin{thebibliography}{10}
\providecommand{\url}[1]{#1}
\csname url@samestyle\endcsname
\providecommand{\newblock}{\relax}
\providecommand{\bibinfo}[2]{#2}
\providecommand{\BIBentrySTDinterwordspacing}{\spaceskip=0pt\relax}
\providecommand{\BIBentryALTinterwordstretchfactor}{4}
\providecommand{\BIBentryALTinterwordspacing}{\spaceskip=\fontdimen2\font plus
\BIBentryALTinterwordstretchfactor\fontdimen3\font minus
  \fontdimen4\font\relax}
\providecommand{\BIBforeignlanguage}[2]{{%
\expandafter\ifx\csname l@#1\endcsname\relax
\typeout{** WARNING: IEEEtran.bst: No hyphenation pattern has been}%
\typeout{** loaded for the language `#1'. Using the pattern for}%
\typeout{** the default language instead.}%
\else
\language=\csname l@#1\endcsname
\fi
#2}}
\providecommand{\BIBdecl}{\relax}
\BIBdecl

\bibitem{standard2018j3016}
S.~Standard, ``J3016-taxonomy and definitions for terms related to driving
  automation systems for on-road motor vehicles,'' 2018.

\bibitem{koopman2016challenges}
P.~Koopman and M.~Wagner, ``Challenges in autonomous vehicle testing and
  validation,'' \emph{SAE International Journal of Transportation Safety},
  vol.~4, no.~1, pp. 15--24, 2016.

\bibitem{xinxin2020csg}
Z.~Xinxin, L.~Fei, and W.~Xiangbin, ``Csg: Critical scenario generation from
  real traffic accidents,'' in \emph{2020 IEEE Intelligent Vehicles Symposium
  (IV)}.\hskip 1em plus 0.5em minus 0.4em\relax IEEE, pp. 1330--1336.

\bibitem{putz2017system}
A.~P{\"u}tz, A.~Zlocki, J.~Bock, and L.~Eckstein, ``System validation of highly
  automated vehicles with a database of relevant traffic scenarios,''
  \emph{situations}, vol.~1, pp. 19--22, 2017.

\bibitem{elrofai2016scenario}
H.~Elrofai, D.~Worm, and O.~O. den Camp, ``Scenario identification for
  validation of automated driving functions,'' in \emph{Advanced Microsystems
  for Automotive Applications 2016}.\hskip 1em plus 0.5em minus 0.4em\relax
  Springer, 2016, pp. 153--163.

\bibitem{amersbach2019functional}
C.~Amersbach and H.~Winner, ``Functional decomposition—a contribution to
  overcome the parameter space explosion during validation of highly automated
  driving,'' \emph{Traffic injury prevention}, vol.~20, no. sup1, pp. S52--S57,
  2019.

\bibitem{ponn2020identification}
T.~Ponn, M.~Breitfu{\ss}, X.~Yu, and F.~Diermeyer, ``Identification of
  challenging highway-scenarios for the safety validation of automated vehicles
  based on real driving data,'' in \emph{2020 Fifteenth International
  Conference on Ecological Vehicles and Renewable Energies (EVER)}.\hskip 1em
  plus 0.5em minus 0.4em\relax IEEE, 2020, pp. 1--10.

\bibitem{menzel2018scenarios}
T.~Menzel, G.~Bagschik, and M.~Maurer, ``Scenarios for development, test and
  validation of automated vehicles,'' in \emph{2018 IEEE Intelligent Vehicles
  Symposium (IV)}.\hskip 1em plus 0.5em minus 0.4em\relax IEEE, 2018, pp.
  1821--1827.

\bibitem{riedmaier2020survey}
S.~Riedmaier, T.~Ponn, D.~Ludwig, B.~Schick, and F.~Diermeyer, ``Survey on
  scenario-based safety assessment of automated vehicles,'' \emph{IEEE access},
  vol.~8, pp. 87\,456--87\,477, 2020.

\bibitem{koren2018adaptive}
M.~Koren, S.~Alsaif, R.~Lee, and M.~J. Kochenderfer, ``Adaptive stress testing
  for autonomous vehicles,'' in \emph{2018 IEEE Intelligent Vehicles Symposium
  (IV)}.\hskip 1em plus 0.5em minus 0.4em\relax IEEE, 2018, pp. 1--7.

\bibitem{liu2021reinforcement}
Y.~Liu, Q.~Zhang, and D.~Zhao, ``A reinforcement learning benchmark for
  autonomous driving in intersection scenarios,'' in \emph{2021 IEEE Symposium
  Series on Computational Intelligence (SSCI)}.\hskip 1em plus 0.5em minus
  0.4em\relax IEEE, 2021, pp. 1--8.

\bibitem{mobileye_rss}
\BIBentryALTinterwordspacing
S.~Shalev{-}Shwartz, S.~Shammah, and A.~Shashua, ``On a formal model of safe
  and scalable self-driving cars,'' \emph{CoRR}, vol. abs/1708.06374, 2017.
  [Online]. Available: \url{http://arxiv.org/abs/1708.06374}
\BIBentrySTDinterwordspacing

\bibitem{zhao2016accelerated}
D.~Zhao, H.~Lam, H.~Peng, S.~Bao, D.~J. LeBlanc, K.~Nobukawa, and C.~S. Pan,
  ``Accelerated evaluation of automated vehicles safety in lane-change
  scenarios based on importance sampling techniques,'' \emph{IEEE transactions
  on intelligent transportation systems}, vol.~18, no.~3, pp. 595--607, 2016.

\bibitem{tuncali2019requirements}
C.~E. Tuncali, G.~Fainekos, D.~Prokhorov, H.~Ito, and J.~Kapinski,
  ``Requirements-driven test generation for autonomous vehicles with machine
  learning components,'' \emph{IEEE Transactions on Intelligent Vehicles},
  vol.~5, no.~2, pp. 265--280, 2019.

\bibitem{de2017assessment}
E.~de~Gelder and J.-P. Paardekooper, ``Assessment of automated driving systems
  using real-life scenarios,'' in \emph{2017 IEEE Intelligent Vehicles
  Symposium (IV)}.\hskip 1em plus 0.5em minus 0.4em\relax IEEE, 2017, pp.
  589--594.

\bibitem{zhao2017accelerated}
D.~Zhao, X.~Huang, H.~Peng, H.~Lam, and D.~J. LeBlanc, ``Accelerated evaluation
  of automated vehicles in car-following maneuvers,'' \emph{IEEE Transactions
  on Intelligent Transportation Systems}, vol.~19, no.~3, pp. 733--744, 2017.

\bibitem{gangopadhyay2019identification}
B.~Gangopadhyay, S.~Khastgir, S.~Dey, P.~Dasgupta, G.~Montana, and P.~Jennings,
  ``Identification of test cases for automated driving systems using bayesian
  optimization,'' in \emph{2019 IEEE Intelligent Transportation Systems
  Conference (ITSC)}.\hskip 1em plus 0.5em minus 0.4em\relax IEEE, 2019, pp.
  1961--1967.

\bibitem{althoff2018automatic}
M.~Althoff and S.~Lutz, ``Automatic generation of safety-critical test
  scenarios for collision avoidance of road vehicles,'' in \emph{2018 IEEE
  Intelligent Vehicles Symposium (IV)}.\hskip 1em plus 0.5em minus 0.4em\relax
  IEEE, 2018, pp. 1326--1333.

\bibitem{ding2020learning}
W.~Ding, B.~Chen, M.~Xu, and D.~Zhao, ``Learning to collide: An adaptive
  safety-critical scenarios generating method,'' in \emph{2020 IEEE/RSJ
  International Conference on Intelligent Robots and Systems (IROS)}.\hskip 1em
  plus 0.5em minus 0.4em\relax IEEE, 2020, pp. 2243--2250.

\bibitem{kuhn2013introduction}
D.~R. Kuhn, R.~N. Kacker, and Y.~Lei, \emph{Introduction to combinatorial
  testing}.\hskip 1em plus 0.5em minus 0.4em\relax CRC press, 2013.

\bibitem{kuhn2004software}
D.~R. Kuhn, D.~R. Wallace, and A.~M. Gallo, ``Software fault interactions and
  implications for software testing,'' \emph{IEEE transactions on software
  engineering}, vol.~30, no.~6, pp. 418--421, 2004.

\bibitem{tuncali2019search}
C.~E. Tuncali, ``Search-based test generation for automated driving systems:
  From perception to control logic,'' Ph.D. dissertation, Arizona State
  University, 2019.

\bibitem{felbinger2019comparing}
H.~Felbinger, F.~Kl{\"u}ck, Y.~Li, M.~Nica, J.~Tao, F.~Wotawa, and
  M.~Zimmermann, ``Comparing two systematic approaches for testing automated
  driving functions,'' in \emph{2019 IEEE International Conference on Connected
  Vehicles and Expo (ICCVE)}.\hskip 1em plus 0.5em minus 0.4em\relax IEEE,
  2019, pp. 1--6.

\bibitem{aerts2018temporal}
A.~Aerts, B.~T. Minh, M.~R. Mousavi, and M.~A. Reniers, ``Temporal logic
  falsification of cyber-physical systems: An input-signal-space optimization
  approach,'' in \emph{2018 IEEE International Conference on Software Testing,
  Verification and Validation Workshops (ICSTW)}.\hskip 1em plus 0.5em minus
  0.4em\relax IEEE, 2018, pp. 214--223.

\bibitem{nas}
B.~Zoph and Q.~V. Le, ``Neural architecture search with reinforcement
  learning,'' \emph{arXiv preprint arXiv:1611.01578}, 2016.

\bibitem{xu2020worst}
L.~Xu, C.~Zhang, Y.~Liu, L.~Wang, and L.~Li, ``Worst perception scenario search
  for autonomous driving,'' in \emph{2020 IEEE Intelligent Vehicles Symposium
  (IV)}.\hskip 1em plus 0.5em minus 0.4em\relax IEEE, 2020, pp. 1702--1707.

\bibitem{sutton2018reinforcement}
R.~S. Sutton and A.~G. Barto, \emph{Reinforcement learning: An
  introduction}.\hskip 1em plus 0.5em minus 0.4em\relax MIT press, 2018.

\bibitem{levine2017cs}
\BIBentryALTinterwordspacing
S.~Levine, ``Cs 294: Deep reinforcement learning,'' 2017. [Online]. Available:
  \url{\url{http://rail.eecs.berkeley.edu/deeprlcourse-fa17/}}
\BIBentrySTDinterwordspacing

\bibitem{carla_simulator}
A.~Dosovitskiy, G.~Ros, F.~Codevilla, A.~Lopez, and V.~Koltun, ``{CARLA}: {An}
  open urban driving simulator,'' in \emph{Proceedings of the 1st Annual
  Conference on Robot Learning}, 2017, pp. 1--16.

\end{thebibliography}

\end{document}